# An Interpretable Object Detection-Based Model for the Diagnosis of Neonatal Lung Diseases using Ultrasound Images

Rodina Bassiouny[1], Adel Mohamed[2], Karthi Umapathy[1], Naimul Khan[1] [*]

[1]*Ryerson University, Toronto, Canada*
[2]*Mount Sinai Hospital, University of Toronto, Toronto, Canada*

*Abstract*— Over the last few decades, Lung Ultrasound (LUS) has been increasingly used to diagnose and monitor different lung diseases in neonates. It is a noninvasive tool that allows a fast bedside examination while minimally handling the neonate. Acquiring a LUS scan is easy, but understanding the artifacts concerned with each respiratory disease is challenging. Mixed artifact patterns found in different respiratory diseases may limit LUS readability by the operator. While machine learning (ML), especially deep learning can assist in automated analysis, simply feeding the ultrasound images to an ML model for diagnosis is not enough to earn the trust of medical professionals. The algorithm should output LUS features that are familiar to the operator instead. Therefore, in this paper we present a unique approach for extracting seven meaningful LUS features that can be easily associated with a specific pathological lung condition: Normal pleura, irregular pleura, thick pleura, A-lines, Coalescent B-lines, Separate B-lines and Consolidations. These artifacts can lead to early prediction of infants developing later respiratory distress symptoms. A single multi-class region proposal-based object detection model faster-RCNN (*f*RCNN) was trained on lower posterior lung ultrasound videos to detect these LUS features which are further linked to four common neonatal diseases. Our results show that *f*RCNN surpasses single stage models such as RetinaNet and can successfully detect the aforementioned LUS features with a mean average precision of 86.4%. Instead of a fully automatic diagnosis from images without any interpretability, detection of such LUS features leave the ultimate control of diagnosis to the clinician, which can result in a more trustworthy intelligent system.

## I. INTRODUCTION

Lung Ultrasound has become a prominent diagnostic tool for assessment and decision-making in care of the neonates with respiratory disease. Several studies found that LUS has many advantages over traditionally used chest radiograph and computed tomography due to its low cost, portability, and safety because of its non-ionizing property [1]. Despite, the ease of acquiring an optimal US image, interpretation of the LUS artifacts remain a challenge due to abundance of artifacts and noise. These artifacts vary in brightness, position, and size. Lack of clear understanding of each LUS artifact may lead to failure of making right diagnosis and consequently inability to provide appropriate therapy. Currently, few trained neonatologists can reliably detect specific features in a LUS scan, a barrier for the wide application of a very beneficial diagnostic imaging modality. Therefore, developing an intelligent system that can assist the physicians to reach an accurate diagnosis of lung via correct interpretation of different LUS features is essential.

Babies born before 37 weeks of gestation constitute 8% of total pregnancies in Canada [2]. One of the common challenges that preterm infants face at delivery is respiratory distress secondary to their lung immaturity or delayed lung transition. Making the right diagnosis of LUS artifacts and linking it to a specific lung pathology is a step towards making the right clinical decision on how to manage these babies. Therefore, incorporating machine learning to detect these artifacts is necessary in guiding the caregivers to reach a decision faster than the conventional ways.

Deep learning and image analysis techniques have shown a huge potential in the medical field. They include classification, segmentation, detection, biometric measurement and quality assessment. However, after extensive discussions with neonatologists, who are expert in neonatal LUS, we determined that to build a trustworthy and interpretable model, simply image classification from ultrasound videos will not be enough to gain their trust. This is due to its non-interpretability and several external factors that affect the final diagnosis such as the baby's gestational age, respiratory symptoms, current respiratory support at the time of LUS scan, past medical history etc.

Previous attempts to detect LUS features were based on an image classification model. A real time blood vessel detection model used AlexNet to determine the position and the size of the vessel in an US image [3]. Another work utilized CNN for the detection of heterogenous thyroid nodules in ultrasound scans [4]. Nerve localization in ultrasound images during delivery of anesthesia has been addressed in [5] using deep CNN with spatio-temporal consistency. In [6], B-lines were detected in lung ultrasound using local maxima in the Radon transform domain. Automatic detection and quantitative scoring of B lines was presented in [7].

To our knowledge, characterizing LUS scan using object detection model on humans has not been done before. As described before, identifying LUS features so it can be linked to a specific diagnosis can lead us to a more trustworthy system rather than a simple image classification model. We identified seven LUS features using a two-stage detector

---

called faster Region Proposal Network fRCNN [8]. The target of our proposed model is to detect the features, while the final diagnosis is done by the caregiver. Qualitative discussions with the caregivers have revealed that this type of interpretable system that aligns with their workflow is more preferable as opposed to a model that simply provides the final diagnosis without any interpretation.

Prior to building an interpretable system, an extensive study of the LUS features and their association to different neonatal lung diseases was done. After several meetings with the LUS expert neonatologist (co-author A.M.), we have mimicked the strategical approach of clinicians to diagnosing five lung conditions based on the occurrence of interpretable features as shown in Fig.1. The five lung conditions are Normal lungs, Respiratory Distress Syndrome (RDS) [9], Transient Tachypnea (TTN) [10], Hemodynamic Patent Ductus Arteriosus (PDA) [11] and Chronic Lung Diseases (CLD) [12]. Clinicians opt for the pleural line as the first LUS feature to look for, before any other feature. Their presence or absence would lead us to recognize other features correlated to different diseases. For instance, TTN occurs only when separate B-lines and pleural sliding exist. Whereas, lungs are considered normal when A-lines accompanied with regular sliding pleurae exist. Compact B line is an overlapping feature between RDS and PDA, with age of patient being the differentiating external factor. Clinician taking care of such patient has to incorporate LUS features, patient's age, and other clinical information to make final diagnosis.

These features are meaningful and important indicators for the type of neonatal lung disease.

The main contributions of this work are summarized as follows:

i. We utilize a two-stage multi-class object detection model fRCNN for detecting seven LUS features in neonatal lung scans: A-lines, normal pleura, irregular pleura, thick pleura, coalescent B-lines, separate B-lines and consolidations. We also implement a one-stage detector model RetinaNet [14] for comparison with fRCNN.
ii. We address the occurrence of several types of Pleura (*Thick+Irregular*) that serve as early predictors to chronic lung pathologies as advised by clinicians. To our knowledge, these anomalies have not been detected with ML before.
iii. We perform experiments on 5 LUS video scans of neonates collected at the Mount Sinai Hospital and professionally annotated by lung ultrasound expert (A.M.).

The rest of the paper is organized as follows: Section II describes the seven ultrasound features that were detected by our proposed model. Ultrasound data mining and annotation details are presented in Section III. The object detection models along with the evaluation metrics are described in Section IV. Performance of the selected models over the newly created dataset is shown in section V and finally, Section VI presents the conclusion of this work.

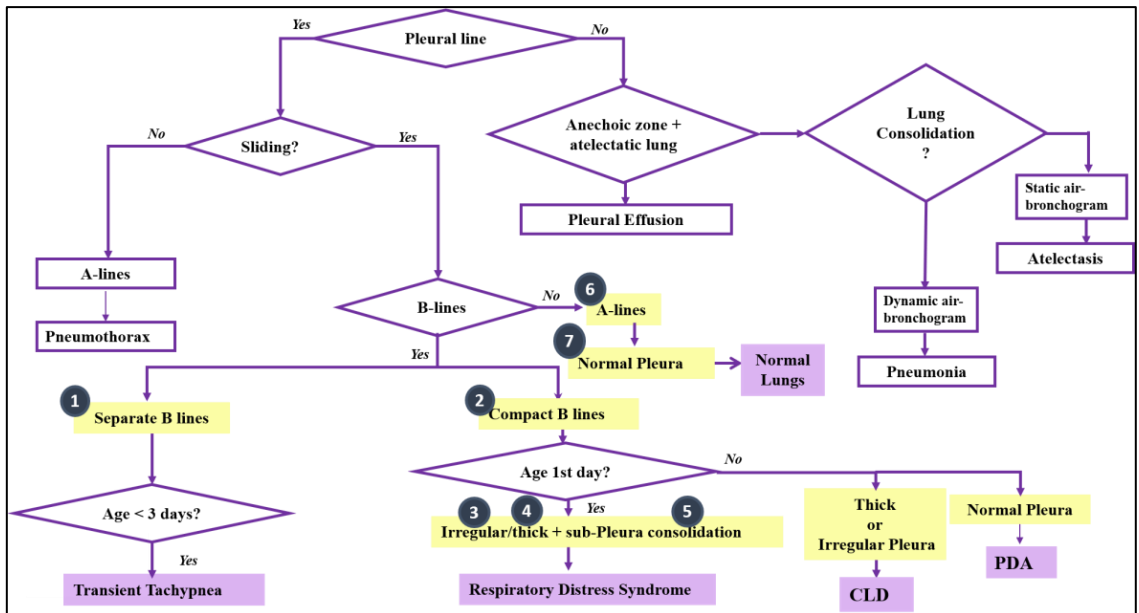

Figure 1: Strategical approach of clinicians in detecting five lung conditions by the occurrence of seven lung features named as A-lines, Thick Pleura, Normal Pleura, Irregular Pleura, Separate B-lines, Coalescent B lines and sub-pleural consolidations.

In [13], an ultrasound model was presented that detects lung abnormalities using Single Shot Detection (SSD) model. They used six separate networks for the detection of six LUS artifacts, hence a more complex model. Our proposed model is a single seven-class fRCNN and has been tested on human LUS rather than swine as in [13]. To the best of our knowledge, types of pleura such as thick pleura or irregular pleura have not yet been addressed by other researchers.

## II- LUNG ULTRASOUND FEATURES IN NEONATAL ULTRASOUND IMAGES

### A. Pleural line and its clinical meaning

Pleurae is composed of 2 layers; a thin membrane (Visceral pleura) that covers the entire lungs and another layer attached to the chest wall (Parietal pleura). In between the two pleurae, there is a thin layer of fluid that allows both layers to slide smoothly during respiration. The interface between pleural fluid and lung tissue, makes this pleural fluid appear very bright (echogenic) in the LUS scan and is considered a true anatomical structure and not a LUS artifact. Normal echogenic pleural line rules out many respiratory diseases. Normally its thickness measures up to 2mm [15], if exceeded then it indicates an underlying etiology and is called *Thick Pleura*. Another abnormal pleura feature in LUS is the *Irregular pleura* which occurs in neonatal CLD as shown in Fig. 2.

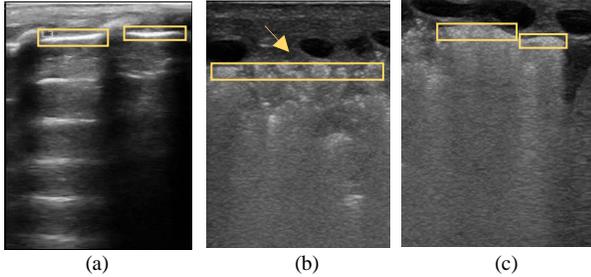

(a)          (b)          (c)

Figure 2: Types of Pleural Line: a) Normal Pleura, b) Irregular Pleura c) Thickened Pleura

*B. B line Artifacts*

B-lines are vertical reverberation image artifacts that fall from the pleural line and extend to the bottom of the screen [16]. Pathological wise, B-lines are corresponding to interstitial and/ or alveolar lung fluid (edema) in the subpleural space and their concentration inversely correlates with lung aeration. For instance, B-lines (<3 lines in one LUS frame) is considered normal as it has no implications on the function of the respiratory system, whereas B lines ≥ 3 is not considered normal. If they are >3mm apart from each other, they are referred to as *separated B lines* whereas B lines ≤3mm apart from one another are called *coalescent B lines*. B-line patterns are shown in Fig. 3.

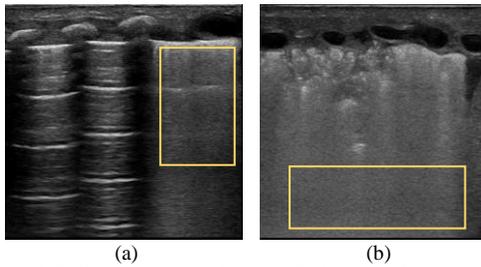

(a)          (b)

Figure 3: B-line patterns (a) Separate B lines, (b) Coalescent B lines

*C. Consolidations*

Lung lobar/hemi-lobar consolidations [17] are hyperechoic tissue-like structure usually visible in a severely flooded lung. It indicates lack of aeration and a progressive building of an inflammatory exudate in the alveoli as shown in Fig. 4.

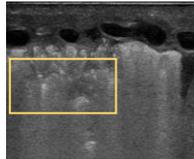

Figure 4: Consolidations

[1] This study was approved by the Mount Sinai and the Ryerson Research Ethics Board.

*D. A-line Artifacts*

A-lines are horizontal reverberation artifacts distanced equally from one another [18]. They mimic the pleural line and are a result of sound waves trapped between the probe and the reflective visceral-pleura interface. An example of A-line artifacts in a Normal lung US scan is shown in Fig. 5.

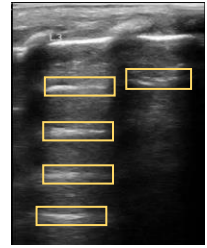

Figure 5: A-line artifacts in a normal LUS scan

III- LUS DATA MINING AND LABELING

*Data extraction*

Lung ultrasound scanning on neonates were performed at Mount Sinai Hospital[1] as per the guidelines for point-of-care lung ultrasound. Raw and short lung ultrasound videos (*sonoanatomy*) were captured from normal and diseased lungs using a linear probe longitudinally. Each patient had a recorded medical history and an age accompanied with the study ID. For each lung condition (Normal, RDS, TTN, Hemodynamic PDA and CLD), complete LUS scans from five neonates were taken. Each LUS video was at least 6 seconds duration and was collected at a frame rate of approximately 18 frames per second.

TABLE I: MARKED LUS FEATRUES PER EACH NEONATAL DISEASE

| Study Case | Etiology | Age | LUS features |
|---|---|---|---|
| Respiratory Distress Syndrome | Surfactant Deficiency | 1st 24 hours | -Thickened and/or Irregular Pleura<br>- Subpleural consolidations<br>- Coalescent B lines |
| Transient Tachypnea | Delayed clearance of lung fluid | 1st 24 hours | - Separate B lines<br>- Normal Pleura |
| Normal Lungs | NA | Any | - A- lines<br>- Normal Pleura |
| Chronic Lung Disease | Lungs injured due to prematurity and ventilation machine | >1 day | - Irregular Pleura<br>- Thick Pleura<br>- Subpleural consolidations<br>- Coalescent B lines |
| Patent Ductus Arteriosus | Pulmonary congestion | >1 day | - Smooth Pleura<br>- Irregular pleura only if neonate has CLD as well.<br>- Consolidations<br>- Coalescent B lines |

All lung diseases, used for this work, along with their etiologies and corresponding features are displayed in Table I. Each lung condition was annotated with its corresponding features frame by frame under the guidance of a neonatology expert (A.M.) in lung ultrasound using the "Dark Label" annotation tool [19]. The final database constituted a total of 1114 A-lines, 374 Normal Pleura, 216 Irregular Pleura, 269 Thick Pleura, 236 Coalescent B-lines, 75 Separate B-lines and 227 Consolidations. A schematic diagram of the proposed automated model is shown in Fig.6.

## IV- THE OBJECT DETECTION MODELS

Building an efficient automated Lung ultrasound detection system requires exploitation of the current *state-of-the-art* architectures. The choice to implement any of these algorithms depends on several factors: Reduction in prediction time, Higher performance (mAP) and reduced complexity. While high performance is prioritized, reasonable prediction time is also important for practical usage. For our Ultrasound dataset, a one stage detector *RetinaNet* and a two stage detector *fRCNN* were implemented and evaluated based on their overall mean average precision and prediction time.

*Anchor Ratio:* Anchors are set to be used as reference when predicting objects location in the image. They were randomly sampled in a mini batch of 256 to maintain a balanced background and foreground. Each LUS artifact was given an multiple anchor ratios corresponding to the coordinates of its location as shown in Table II. These ratios are calculated as the bounding box height ($Y_{max}-Y_{min}$) to its width ($X_{max}-X_{min}$). Anchors eliminate the need of defining a variable length bounding boxes on the raw images and use ratios and scales to fit variable sized features.

During training, the model learns to predict offsets from ground truth boxes and further uses them to predict the raw bounding box coordinates on the reshaped images.

*Anchor Scale:* LUS features of larger area such as Separate B lines and Coalescent B lines were given a higher anchor scale of 256 and 128 whereas the rest of the smaller pathologies were given 32 and 64.

*Anchor Size:* Number of anchors *k* set per spatial position in a LUS is 48 according to (1). Total potential anchors per image is defined based on the generated feature map.

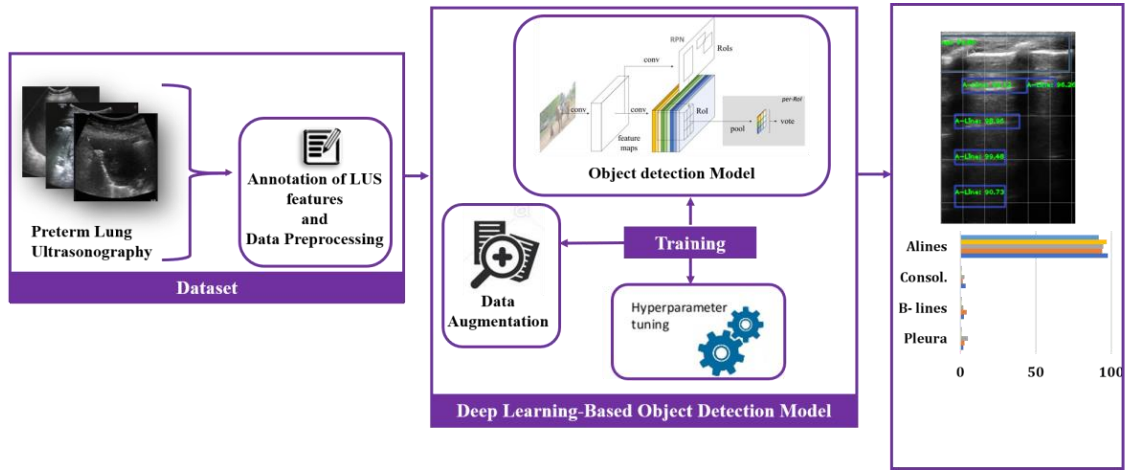

Figure 6: Schematic diagram of the proposed automated model

### A. Faster RCNN

Faster RCNN is a two-stage object detector that utilizes features maps obtained from a pre-trained neural network and generates region proposals using Region Proposal Network *(RPN)*. Predicted regions are reshaped and fed into a pooling layer before passing it into an RCNN. A schematic diagram of the *f*RCNN model is shown in Fig. 7.

*Backbone:* There are several pre-trained Image classification Network architectures that can be used as a base network for *f*RCNN. ResNet-152 has 152 layers making around 60 Million parameters whereas MobileNet is a smaller network with approximately 3.3 Million parameters. VGG-16 was opted for this ultrasound project due to its optimized speed, smallest architecture and lightweight factor if used as an inference, on a device in future premises. Feature maps are extracted from (*block5_pool*) layer of dimension [7,7,512]. They encode all the information for the Ultrasound image while maintaining the location of each LUS feature relative to the original image.

For instance, feature maps extracted from RDS scans had a dimension of 467x300x1, so their potential anchors were 6,724,800 according to (2).

Number of anchors per spatial Position = Anchor_Scale *Anchor_Ratio (1)

Potential Anchors = $conv_{width} \times conv_{height} \times conv_{depth}$ x Anchors per point (2)

TABLE II: ANCHORS RATIO PER EACH LUS FEATURE

| LUS features | $X_{MIN}$ | $Y_{MIN}$ | $X_{MAX}$ | $Y_{MAX}$ | Anchor Ratios |
|---|---|---|---|---|---|
| A-lines | 49 | 149 | 168 | 180 | 31/119 |
| Normal Pleura | 53 | 60 | 174 | 88 | 28/121, |
|  | 38 | 71 | 138 | 97 | 26/100 |
| Thick Pleura | 416 | 94 | 484 | 121 | 27/68 |
| Irregular Pleura | 191 | 53 | 365 | 80 | 27/175, |
|  | 182 | 46 | 441 | 72 | 26/259 |
| Separate B-lines | 269 | 82 | 444 | 450 | 368/175 |
| Coalescent B-lines | 26<br>108 | 313<br>201 | 468<br>410 | 466<br>325 | 153/442,<br>124/302 |
| Consolidations | 214<br>6<br>190 | 79<br>114<br>84 | 433<br>308<br>363 | 156<br>247<br>155 | 77/219,<br>133/302,<br>71/173 |

truth values. Smooth L1 loss is suggested for the regression loss [8]. The model was trained over 100 epochs with epoch step size of 100, a stochastic gradient of 0.9, learning rate of e$^{-05}$. A 12GB NVIDIA Tesla K80 GPU was utilized for training the fRCNN. The backbone network was a pretrained VGG-16 on ImageNet. During testing, a bounding box threshold of 0.8 and non-maximum suppression of 0.2 were set.

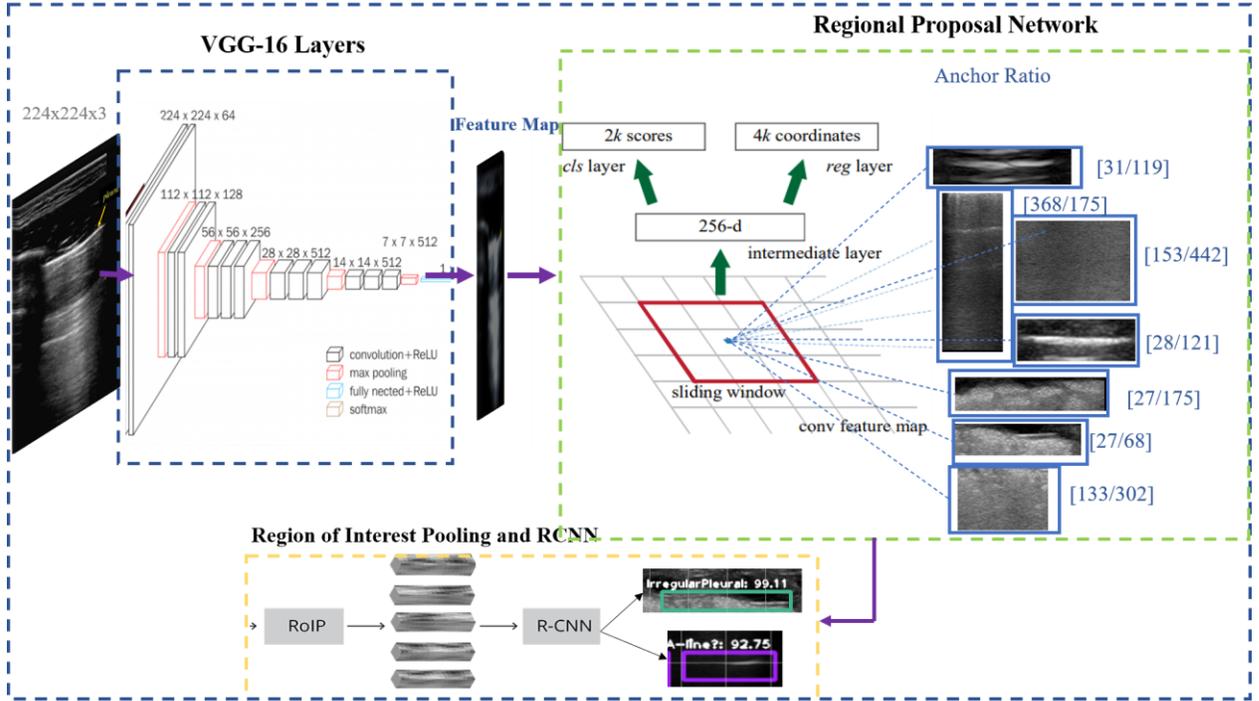

Figure 7: Schematic diagram of the faster RCNN model

*Region Proposal Network (RPN):* RPN is a convolutional network that takes in feature maps of any input size and generates a good set of proposals. Using the anchors mentioned in Table II, the convolutional network generates two outputs per anchor: Objectness score and bounding box regression. The objectness score is referred to the probability of it being an object or not. The second output is used for adjusting the anchors to better fit each LUS feature. The classification layer has 2x*k* channels and the regression has 4x*k* channels, where k is the number of anchors per spatial position.

*ROI Pooling:* Each proposal is projected onto a fixed feature map of [7x7x512]. This is to fix the variable sized input feature maps. No class is yet assigned to them and so they are passed to RCNN.

*Training and Testing Parameters for fRCNN:* All anchors are taken during the training stage. Anchors are matched to the ground truth boxes with Jaccard overlapping index of 0.5. Values above that threshold were considered foreground and the rest were background. For the detection of the 7 LUS features, twelve anchor ratios were used as shown in Table II. Anchor scales were set as 128, 256, 64 and 32. The main objective of the *f*RCNN is learning to minimize the difference between predicted bounding box co-ordinates and the ground

*B. RetinaNet*

As a single stage detector, RetinaNet [14] is known for its high inference speed. This model combines the idea of feature pyramids and a novel loss function called the focal loss. Like *f*RCNN, its structure composes a feature extractor backbone, a feature pyramid network and finally subnets. For this work, a ResNet-18 was chosen as the backbone due to its optimized speed and small architecture. In *f*RCNN, the final feature map from VGG-16 was considered, however, in RetinaNet every feature map generated from each conv layer in Resnet-18 was considered. This saves the representational hierarchy or the Multi-scale Feature Pyramidal shape. Unlike fRCNN, which generates anchors only in the final feature map from VGG-16, RetinaNet generates multi-level anchors per the default conv layers in the backbone.

*Training and Testing Parameters for RetinaNet*

Anchor ratios were set as per Table II. Anchor scales were set as [0.25, 2,5]. Potential anchors per spatial point were 36. Using early stopping, the model was trained over 58 epochs with a learning rate of 3.64 e$^{-4}$. During testing, a bounding box threshold of 0.8 and a non-maximum suppression of 0.1 was set.

*C. Evaluation Metrics*

Mean Average Precision was selected to evaluate the chosen

TABLE III: TRAINING STATISTICS AND TESTING PERFORMANCE.

| Lung Pathologies | fRCNN | | | RetinaNet |
|---|---|---|---|---|
| | mAP | | | |
| | IoU=0.4 | IoU=0.45 | IoU=0.5 | mAP % |
| A-lines | 81.9 | 74.8 | 67.5 | 30.55 |
| Normal Pleura | 89.3 | 87.14 | 85.4 | 25.6 |
| Thick Pleura | 52 | 51.36 | 51.36 | 56.5 |
| Irregular Pleura | 84.93 | 74.69 | 68 | 64.8 |
| Consolidations | 97.5 | 95.6 | 91.8 | 89.10 |
| Separate B-lines | 99.9 | 98.99 | 91.5 | 99 |
| Coalescent B lines | 99.3 | 97.99 | 93.16 | 18.4 |
| Total mAP | 86.4% | 82.93% | 78.38% | 54.85% |
| Prediction Time Per frame | 12.5s | | | 2.5s |
| Time per epoch | 31s | | | 10s |

TABLE IV: DATASET TRAIN/TEST SPLIT

| Neonatal Lung Conditions | Training LUS frames | Testing LUS frames |
|---|---|---|
| Respiratory Distress Syndrome | 63 | 12 |
| Transient Tachypnea | 67 | 8 |
| Patent Ductus Arteriosus | 65 | 10 |
| Chronic Lung Disease | 77 | 8 |
| Normal | 101 | 11 |

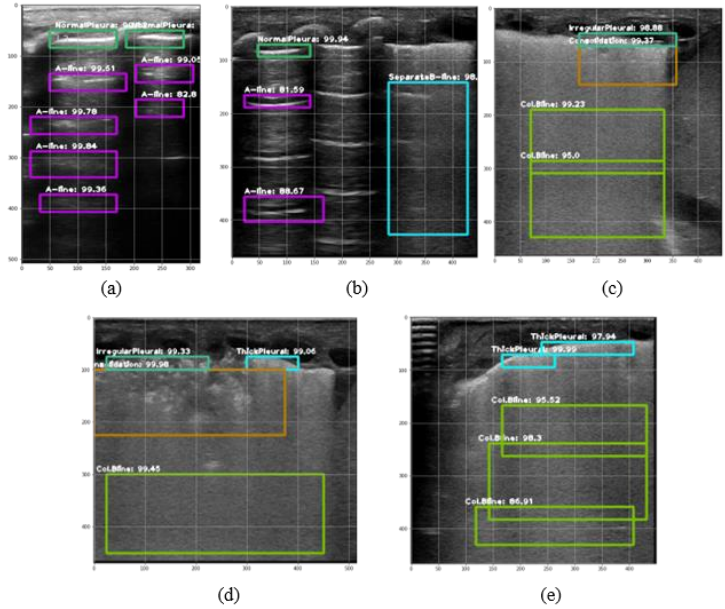

Figure 8: Sample results for Multi-object fRCNN model. Features associated with each neonatal disease are highlighted along with their confidence scores: (a) Normal LUS scan with its detected features: A-lines and Normal Pleurae, (b) Transient Tachypnea scan with detected A-lines, Normal Pleura and Separate B-lines. (c) Respiratory distress syndrome associated with detected Irregular pleura, subpleural consolidations and Coalescent B lines, (d) Chronic Lung Disease scan with detected Irregular pleura, subpleural consolidations and Coalescent B lines in one lung side. Detected thick pleura has been detected in the other lung side, (e) Patent Ductus Arteriosus LUS Scan with its associated detected features: Coalescent B lines and thick pleura.

object detection models. Each model outputs the bounding box locations accompanied by class probability per each LUS scan. Subsequently, decisions are made based on two factors: Probability score threshold and Jaccard Index. In this work, mAP was calculated per each LUS feature under several Jaccard indexes: 0.4, 0.45, 0.5. Consider two bounding boxes C1 and C2, the Intersection over Union $IoU$ will be

$$\text{IoU} = \frac{|C1 \cap C2|}{|C1|+|C2|-|C1 \cap C2|} \quad (3)$$

After training, bounding boxes with a certain selected IoU are localized per image. Recall and Precision are then calculated as in (4) and (5),

$$R = \frac{\text{TP}}{\text{TP+FN}} \quad (4) \qquad P = \frac{\text{TP}}{\text{TP+FP}} \quad (5)$$

where TP, FN and FP are the true positives, false negatives and false positives respectively.

Mean Average Precision is used to evaluate object detection models. It compares detected boxes with the ground truth boxes and returns a score. The higher the value, the more accurate is the localization property of the model. Average precision itself is the measured area under the precision recall curve as shown in (6).

$$AP = \int_0^1 P(R)dR \qquad (6)$$

AP is computed per each LUS feature in each frame and averaged across all frames to obtain mAP per LUS feature. Lastly, mAP of all LUS features are averaged to obtain the mAP of the model.

## V- RESULTS

All results of the evaluation are summarized in Table III. The models were trained on a 70%/30% for train/validation dataset as shown in Table IV. The split was random and cross validation was not performed, since it is not a common practice for object detection due to the localized nature of detection. The dataset is highly imbalanced with A-line as the most abundant feature with 1114 and Separate B lines as the least of only 75. Parameters such as mAP per LUS feature, prediction time (speed) and time per epoch were considered as evaluation parameters for the best-selected model. The fRCNN model was evaluated under several IoUs: 0.4, 0.45 and 0.5. In terms of mAP, fRCNN could capture six out of seven features with an accuracy of above 80% and only one feature below 60%. On the other hand, RetinaNet captured two out of seven lung features with mAp above 80% whereas three LUS features were below 40%. Both models accurately localized Separate B-lines with a mAP of above 95%. Overall mAP for fRCNN was 86.4% (*IoU=0.4*) as compared to only 54.85% achieved by RetinaNet. This is likely due to the extra network structure in fRCNN that proposes the regions before passing them onto RCNN, thus improving its localization attribute. Next factor for the model's evaluation was its speed in prediction of features per frame. RetinaNet was approximately five times faster than fRCNN. Finally, the calculated training time per epoch for fRCNN was about three times higher than RetinaNet. This was due to the higher number of stages, making it a more time consuming model. However, the much higher mAP of fRCNN makes it an appropriate choice despite slower speed. Sample frames from the five LUS videos with their corresponding detected features are displayed in Fig. 8.

## VI- CONCLUSION

In summary, we have shown that the selected object detection model has achieved a high overall mAP of 86.4%. The model was able to capture seven LUS features that are associated with different neonatal lung diseases. We were able to implement an end-to-end system that coincides with the workflow of the neonatologists' strategy in diagnosing a respiratory disease and interpreting the LUS scans as preferred and desired by the caregivers.

In the future, we are hoping to receive more data and expand our database by including other lung pathologies such as Pneumonia and pleural effusion. We are aiming to extend our work on several other Lung ultrasound views as well. This will help validate our proposed method.

ACKNOWLEDGMENT

NSERC's funding through an Alliance grant is appreciated. We thank Jenna Ibrahim from Mount Sinai Hospital for coordinating the data collection process.